%% file: wacv_final.tex
\documentclass[10pt,twocolumn,letterpaper]{article}

\usepackage{wacv}

\title{OX-NeRF: 3D X-ray Tomography Reconstruction from Sparse Views Using Implicit Neural Representation}

\author{Thomas Welsch\\
Department of Engineering Science\\
University of Oxford\\
Oxford, United Kingdom
\and
Min-Hsin Tu\\
Department of Engineering Science\\
University of Oxford\\
Oxford, United Kingdom
\and
David J. Chapman\\
Department of Engineering Science\\
University of Oxford\\
Oxford, United Kingdom
\and
Daniel E. Eakins\\
Department of Engineering Science\\
University of Oxford\\
Oxford, United Kingdom
}

\input{wacv_packages}

\definecolor{wacvblue}{rgb}{0.21,0.49,0.74}
\usepackage[breaklinks,colorlinks,allcolors=wacvblue]{hyperref}
\usepackage[capitalize]{cleveref}

\def\wacvPaperID{1362}
\def\confName{WACV}
\def\confYear{2027}

\begin{document}

\maketitle

\input{front/Abstract}

\input{contents/Introduction}

\input{contents/Related_Work}

\input{contents/Preliminary}
\input{contents/Method}

\input{contents/Experiments}

\input{contents/Conclusions}
\input{back/Acknowledgement}

\clearpage
{
    \small
    \bibliographystyle{ieeenat_fullname}
    \bibliography{back/references}
}

\end{document}

%% file: wacv_packages.tex
\usepackage{amssymb}
\usepackage{amsthm}
\usepackage{textcomp}
\usepackage{booktabs}
\usepackage{multirow}
\usepackage{makecell}
\usepackage{adjustbox}
\usepackage{array}
\usepackage{float}
\usepackage{stfloats}
\usepackage{colortbl}
\usepackage[ruled,vlined,linesnumbered]{algorithm2e}

%% file: front/Abstract.tex
\begin{abstract}

NeRF and Gaussian splatting methods have been successfully applied on  X-ray scenes where the views are too sparse for 3D reconstruction via classical methods. Ultra-sparse scenes with 10 or fewer views such as those with high-rate or low-dose acquisition still, however, present a significant challenge. To address this problem we present a new framework, Optimised X-ray Neural Radiance Fields (OX-NeRF), that combines cross-scene feature learning with scene-specific optimisation to reconstruct sets of related scenes. OX-NeRF employs a convolutional neural network (CNN) to identify cross-scene features while maintaining scene-specific multi-resolution hash grids of spatial features. The paired representations are fused and passed to a multilayer perceptron (MLP); the CNN, hash grids and MLP are then jointly optimised end-to-end. Benchmarking on parallel-beam and cone-beam X-ray datasets shows OX-NeRF provides significantly higher reconstruction accuracy on ultra-sparse scenes compared to existing radiance field methods.
\end{abstract}

%% file: contents/Introduction.tex
\section{Introduction} \label{sec:introduction}

X-ray 3D reconstruction becomes increasingly ill-posed as the number of available projections decreases \cite{Crowther1970TheMicroscopy}, and in many imaging settings that number is severely constrained.
The cumulative radiation dose delivered to a subject can impose a practical limit on how many projections may be acquired \cite{Brenner2007ComputedExposure,Kiss2024BenchmarkingTasks}.
High-rate acquisition imposes a different limit: when the object evolves or is destroyed within microseconds, there is no opportunity to rotate it, and only the projections captured simultaneously are available \cite{Vagovic2018HardApproaches,Asimakopoulou2024DevelopmentImaging}.
Both constraints can lead to an ultra-sparse regime in which a scene is observed by ten or fewer projections \cite{Vagovic2018HardApproaches}.
At this level of sparsity the null space of the inverse problem is large enough that classical reconstruction cannot be reliably applied, and alternative approaches are required \cite{Zhang2023ONIX:Views}.

We propose Optimised X-ray Neural Radiance Fields (OX-NeRF), a framework for X-ray reconstruction designed for this regime.
OX-NeRF takes as input a set of related scenes and trains a single model from which any of them can be reconstructed.
A shared convolutional encoder learns cross-scene features common to the set \cite{Yu2020PixelNeRF:Images}, which let the model resolve internal geometry that cannot be inferred from a single scene's projections alone.
At the same time a multi-resolution hash encoding \cite{Muller2022InstantEncoding} is learned for each scene, so that object-specific structure is stored separately rather than competing for shared capacity.
The two sets of features are fused and passed to an MLP which predicts attenuation.
The full pipeline is optimised by self-supervision with predicted attenuation integrated along each ray and compared against the measured projections.
Once trained, the model can be queried to produce a full 3D reconstruction of any scene in the set.

We evaluate OX-NeRF on four datasets restricted to the ultra-sparse regime of ten
projections or fewer. OX-NeRF achieves the
highest 3D reconstruction accuracy against existing self-supervised radiance
field methods on every dataset at each projection count from four to ten. This
establishes it as the state of the art for ultra-sparse X-ray
reconstruction.

%% file: contents/Related_Work.tex
\section{Related Work} \label{sec:related}

\subsection{Computed Tomography}
Computed tomography (CT) recovers a 3D volume from 2D projections captured
at multiple angles. Classical tomographic reconstruction algorithms fall into analytical and
iterative families. Analytical methods such as filtered back projection
solve the Radon transform and its inverse \cite{Kak2001PrinciplesImaging,
Shepp1974FOURIERSECTION.} and produce a volume almost instantly. The
resolution they can attain is tied to the number of projections by the
Crowther criterion \cite{Crowther1970TheMicroscopy}, which requires roughly
$\pi D / d$ evenly spaced projections to resolve features of size $d$ in
an object of diameter $D$. Far below that count they introduce severe
streak artefacts. Iterative methods
\cite{Andersen1984SimultaneousAlgorithm,
Karczmarz1937AngenaherteGleichungen} formulate reconstruction as an
optimisation problem with a regulariser such as total variation
\cite{Niu2014Sparse-viewRegularization,
Yazdanpanah2016Sparse-viewRegularization}. They suppress artefacts at
moderate sparsity but at ten projections or fewer a hand-designed
regulariser is not sufficient to determine the solution
\cite{Barutcu2021Limited-anglePriors, Engl2015RegularizationProblems,
Pan2009WhyReconstruction}. Supervised deep learning methods replace the
hand-designed regulariser with priors learned from datasets of CT volumes
and use them to inpaint projections, denoise volumes or predict the
reconstruction directly \cite{Corona-Figueroa2022MedNeRF:X-ray,
Guo2023Deep-learningComposites, Kiss2024BenchmarkingTasks,
Wang2022Two-stageTomography, Ying2019X2CT-GAN:Networks,
Zheng2023UltrasparseImaging}. These methods require labelled representative volumetric data to perform well which is impractical to obtain in many imaging settings. 

\subsection{Radiance fields}
Radiance field methods reconstruct a scene by fitting a differentiable
representation of it to a set of images \cite{Tewari2022AdvancesRendering}:
the representation is rendered from each camera, the rendering is compared
with the image, and the representation is updated until the two agree.
NeRF \cite{Mildenhall2021NeRF:Synthesis} represents the scene as a
continuous function of position learned from posed images through volume
rendering. Fourier feature encodings \cite{Tancik2020FourierDomains} let
the network represent fine detail, and learnable multiresolution encodings
\cite{Martel2021ACORN:Representation, Muller2022InstantEncoding} together
with fully fused networks \cite{Muller2021Real-timeTracing} cut training
from hours to seconds. 3D Gaussian Splatting (3DGS)
\cite{Kerbl2023GaussianSplatting} replaces NeRF's implicit function with
explicit Gaussian primitives that are rasterised rather than sampled along
rays. Both NeRF and 3DGS have been adapted to X-ray imaging by replacing
view-dependent colour with attenuation and the compositing step with a
line integral. Because the rendered projection can be compared directly
with the measured one, these representations enable tomographic
reconstruction by self-supervised learning: the representation is
optimised against the measured projections alone and no volumetric ground
truth is required. Neural attenuation fields of this kind have been fitted
to single scans with a range of encodings and sampling strategies
\cite{Fu2025Sparse-ViewFields, Zang2021IntraTomo:Prediction,
Zha2022NAF:Reconstruction, Zhou2024rho-NeRF:Reconstruction}. SAX-NeRF adds
a line-segment transformer and structure-aware sampling to capture fine
internal structure \cite{Cai2024Structure-AwareReconstruction}. On the
explicit side X-Gaussian adapts splatting to X-ray novel view synthesis
\cite{Cai2024RadiativeSynthesis} and R\textsuperscript{2}-Gaussian
rectifies its integration so that a tomographically consistent volume can
be extracted \cite{Zha2024R2Gaussian}. These methods fit one scene at a
time and have not been evaluated in the ultra-sparse domain.

\subsection{Generalisable Radiance Fields}
Generalisable radiance fields train a single model across many scenes so that it
can reconstruct a new scene from its input views without per-scene
optimisation \cite{Zhang2025AdvancesSurvey}. A shared encoder extracts
features from the input views, which are used either to condition a neural
field \cite{Yu2020PixelNeRF:Images} or to predict the parameters of
Gaussian primitives directly \cite{Charatan2024PixelSplat:Reconstruction}.
In X-ray imaging ONIX \cite{Zhang2023ONIX:Views, Zhang20254D-ONIXLearning,
Yue2025Physics-informedData} conditions a neural field in this way under a
parallel-beam renderer and trains across a set of scenes so that the
shared prior compensates for missing angular coverage. Its prior is held
entirely in shared weights, which limits the capacity available to any
single scene. OX-NeRF also features a cross-scene encoder but pairs it
with an explicit per-scene spatial representation that is optimised for
each scene in the set, so that transferable priors and scene-specific
geometry are stored and updated separately.

%% file: contents/Preliminary.tex
\section{Preliminaries} \label{sec:preliminary}

\subsection{X-ray image formation} \label{sec:xray-model}
X-ray image formation is modelled through the standard weak-interaction
approximation. The refractive index $n$ characterises how electromagnetic
radiation interacts with matter, and is written as
\begin{equation}
    n = 1 - \delta + i\beta,
\label{equ:refactive index}
\end{equation}
where $\delta$ describes phase shift and $\beta$ describes attenuation. In
the X-ray regime the refractive index is close to one, which allows a
compact wave-based description of transmission through matter. Under the
projection approximation \cite{Paganin2006CoherentOptics}, the exit wave can
be written as
\begin{equation}
    \psi_{\text{e}}(x, y) = \psi_0(x, y) \cdot \exp\left(-i k \int n(x, y, z) \, dz \right),
\label{equ:x-ray propagation}
\end{equation}
where $\psi_0$ and $\psi_{\text{e}}$ denote the incident and exit waves,
$k$ is the wave number, and the integration is along the propagation
direction $z$.

A projection is the image the transmitted beam leaves on the detector. Each
pixel of that image sits at the end of one ray through the scene, and by
Eq.~\ref{equ:x-ray propagation} the wave arriving at it has accumulated the
refractive index along the whole of that ray. Taking the logarithm of the
transmission at the pixel therefore gives a line integral of $n$ along its
ray, which is the relation between the 3D field and the 2D projection,
\begin{equation}
    I(\mathbf{r}) = \ln\left( \frac{\psi_{\text{e}}}{\psi_0} \right)
    = -i k \int_{t_n}^{t_f} n\left(\mathbf{r}(t)\right) \, dt
    \approx -i k \sum_{j=1}^{N_{\mathrm{s}}} n_j \, \Delta_j,
\label{equ:path integration}
\end{equation}
where $\mathbf{r}(t)$ is the ray of Eq.~\ref{equ:raytracing} traced between
the near and far bounds $t_n$ and $t_f$ of a bounding volume enclosing the
scene, $n_j$ is the field value at the $j$-th of $N_{\mathrm{s}}$ sample
points, and $\Delta_j$ is the spacing between adjacent samples. Unlike
visible-light imaging there is no occlusion. Every point along the ray
contributes in proportion to its attenuation, so a single pixel constrains
the whole line rather than a surface. The projections considered here are
attenuation projections, to which only the $\beta$ term of $n$ contributes.

Eq.~\ref{equ:path integration} runs from 3D to 2D: given the attenuation
$\beta$ throughout the scene, its $\beta$ term yields the attenuation
projection along any ray. Reconstruction is the inverse problem. The aim is
to recover the true field $\beta$ at every point of the scene, of which only
the measured projections are observed. Sampled on a voxel grid, that field
is the 3D reconstruction.

\subsection{Acquisition geometry} \label{sec:geometry}
Scene and detector coordinates are related by the affine transformations of
the acquisition geometry \cite{Szeliski2022ImageFormation}, so every
detector pixel defines a ray
\begin{equation}
    \mathbf{r}(t) = \mathbf{r}_o + t \, \mathbf{r}_d, \quad t \in \mathbb{R},
\label{equ:raytracing}
\end{equation}
with origin $\mathbf{r}_o$ and direction $\mathbf{r}_d$. Parallel-beam and
cone-beam acquisitions differ only in how $\mathbf{r}_o$ and $\mathbf{r}_d$
follow from the recorded geometry, so one renderer serves both without any
change to the model. The same transformations can be run in reverse to map a 3D point $\mathbf{x}$ to its position $\pi_p(\mathbf{x})$ on the detector of any projection $p$.

%% file: contents/Method.tex
\begin{figure*}[!t]
	\centering
    \includegraphics[width=\textwidth]{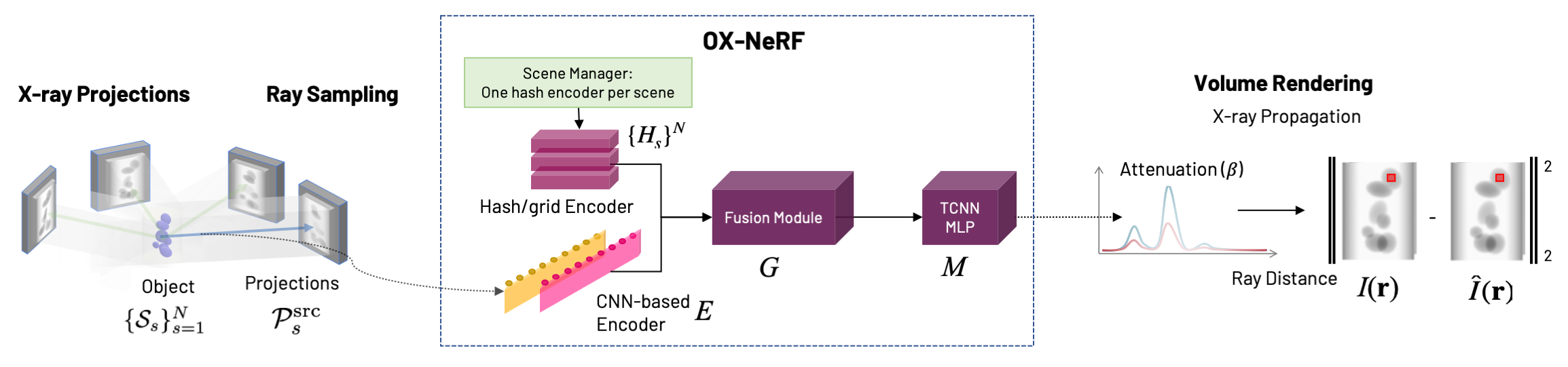}
    \caption{\textbf{Overview of the OX-NeRF pipeline.}
    \emph{Left:} each scene is recorded by a few X-ray projections at known angles, and a batch of rays is drawn from them by the rule of Section~\ref{sec:sampling}.
    \emph{Centre:} the \emph{scene manager} activates the hash grid of the current scene, so the hash encoder reads only that scene's table, while the shared convolutional encoder reads a pixel-aligned descriptor from the source projections; the fusion module combines the two and a fully fused head returns an attenuation value.
    \emph{Right:} attenuation is accumulated along the ray by Eq.~\ref{equ:path integration} and compared with the measurement by mean squared error.
    Only the hash grids are specific to a scene.}
	\label{fig:pipeline}
\end{figure*}

\section{Method} \label{sec:method}

\subsection{Pipeline}
The OX-NeRF pipeline, visualised in Figure~\ref{fig:pipeline}, comprises a convolutional encoder, multi-resolution hash encodings \cite{Muller2022InstantEncoding}, a fusion module and an MLP prediction head.
The pipeline takes as input a set of related scenes recorded under a common acquisition setting. 
Each scene is a number of X-ray projections captured at known angles. 
The pipeline's output is a single trained model from which the full attenuation volume of any scene in the set can be reconstructed. 
The key design choice is the split between shared and per-scene components: the encoder, fusion module and MLP learn features that pertain to the whole set, while a hash encoding learns to capture scene-specific features.

\begin{algorithm}[t]
\caption{OX-NeRF Training Loop}
\label{alg:training}

\SetCommentSty{textnormal}
\KwIn{scenes $\{\mathcal{S}_s\}_{s=1}^{N}$ with source projections
$\mathcal{P}^{\mathrm{src}}_s$ and target projections
$\mathcal{P}^{\mathrm{tgt}}_s$}
initialise the encoder $E$, fusion module $G$, prediction head $M$, and
per-scene hash-grid encodings $\{H_s\}_{s=1}^{N}$\;
\For{$\mathrm{epoch} \leftarrow 1$ \KwTo $N_{\mathrm{ep}}$}{
  \If{$\mathrm{epoch} \equiv 1 \pmod{\tau}$}{
    re-render $\mathcal{P}^{\mathrm{tgt}}$ of the next $B$ scenes, cache
    $R \leftarrow |\hat{I} - I|$\tcp*[l]{Sec.~\ref{sec:sampling}}
  }
  \For{$\mathrm{scene}$ $s$ $\mathrm{in}$ $\mathrm{shuffled}$ $\mathrm{order}$}{
    $\{F_p\}_{p \in \mathcal{P}^{\mathrm{src}}_s} \leftarrow
    E(\mathcal{P}^{\mathrm{src}}_s)$\tcp*[l]{encode}
    draw rays from $\mathcal{P}^{\mathrm{tgt}}_s$, $\Pr \propto R +
    \text{floor}$\tcp*[l]{Eq.~\ref{equ:residual sampling}}
    stratified sample $\mathbf{x}$ along each
    ray\tcp*[l]{Eq.~\ref{equ:raytracing}}
    \ForEach{$\mathbf{x}$, $p \in \mathcal{P}^{\mathrm{src}}_s$}{
      $\beta_p \leftarrow
      M\bigl(G\bigl[F_p(\pi_p(\mathbf{x})),\, H_s(\mathbf{x})\bigr]\bigr)$\;
    }
    $\beta(\mathbf{x}) \leftarrow \textstyle\sum_p w_p
    \beta_p$\tcp*[l]{Sec.~\ref{sec:aggregation}}
    $\hat{I} \leftarrow$ integrate $\beta$ along each
    ray\tcp*[l]{Eq.~\ref{equ:path integration}}
    $\mathcal{L} \leftarrow \mathrm{MSE}(\hat{I}, I)$, update $E$, $G$, $M$,
    $H_s$\;
  }
}
\end{algorithm}

OX-NeRF's training loop is described in Algorithm~\ref{alg:training}. 
Each scene $s$ contributes two subsets drawn from its projections: the source projections $\mathcal{P}^{\mathrm{src}}_s$, which the convolutional encoder turns into feature maps, and the target projections $\mathcal{P}^{\mathrm{tgt}}_s$, whose measured pixels the loss is computed against. We write $p$ for a single source projection and $P = |\mathcal{P}^{\mathrm{src}}_s|$ for how many there are, the same count in every scene. 
One iteration operates on a single scene and one epoch covers every scene in the set. 
In an iteration, rays are cast from the pixels of the target projections using a residual-guided strategy: the target projections are periodically re-rendered, and rays are drawn preferentially where the rendered and
measured pixels disagree. 
3D Points are obtained by stratified sampling along each ray where it crosses a fixed bounding volume shared by every scene in the set \cite{Mildenhall2021NeRF:Synthesis}.
Each point $\mathbf{x}$ is then described twice: by the feature map $F_p$ of each source projection $p$, read at the 2D position $\pi_p(\mathbf{x})$ where $\mathbf{x}$ lands on that projection, and by the scene's hash encoding $H_s$, queried at the 3D coordinates of $\mathbf{x}$.
The two descriptions are fused per source projection and mapped to an attenuation value by the MLP, and geometry-aware weighting then combines the per-projection values into the point's attenuation.
Integrating the attenuation along each ray by Eq.~\ref{equ:path integration} renders the pixel, and the mean squared error against the measured target pixels forms the loss.

The loss is backpropagated at every iteration to keep each update scene-specific: gradients from one scene's loss reach the shared components and that scene's hash encoding, and never touch another scene's. 
The scene manager keeps every hash encoding resident in GPU memory and routes each query to the active scene's encoding, so the per-scene binding costs no transfers. 
Over an epoch the shared components accumulate updates from every scene in the set, while each hash encoding is shaped only by feedback from its corresponding scene.

\subsection{Ray Sampling} \label{sec:sampling}
Each training iteration renders only a small batch of rays, so where those rays land matters. 
OX-NeRF selects them using the residual between rendered and measured projections, which indicates where the model is currently weak.
Each target projection keeps an error map $R$, the cached absolute difference
between its rendered and measured pixels, and rays are drawn preferentially
where $R$ is large. The maps are rebuilt from the model as it trains, so the
sampling follows the model's weaknesses as they move: a region that is
repaired stops attracting rays, and the budget shifts to whatever the model
still gets wrong.

The ray budget of an iteration is divided evenly across the target
projections of the active scene, so no single angle dominates an update.
Within each projection, rays are drawn from
\begin{equation}
    \chi(i) = (1-\varepsilon)\,\frac{R_i^{\alpha}}{\sum_{i'=1}^{N_{\mathrm{pix}}} R_{i'}^{\alpha}} \; + \; \varepsilon \, \frac{1}{N_{\mathrm{pix}}},
\label{equ:residual sampling}
\end{equation}
where $R_i$ is the cached error at pixel $i$, $i'$ runs over all
$N_{\mathrm{pix}}$ pixels of that same projection, and the denominator is what
makes the scores a distribution. The exponent $\alpha$ sets how sharply the first term concentrates on the
largest errors, with $\alpha = 1$ making the probability proportional to
the error. The uniform term receives a fraction $\varepsilon$ of the budget
and keeps supervision flowing to regions the model currently renders well,
so they cannot drift between refreshes.

Rebuilding every error map at every step would dominate the cost of training.
Because of this we instead employ a refresh schedule as seen on lines 3 and 4 of Algorithm~\ref{alg:training}:
every $\tau$ epochs the next $B$ scenes in a fixed cyclic order have their
target projections re-rendered and their maps replaced. Between refreshes a
scene keeps its most recent maps, and a scene the cycle has not yet reached
draws its rays uniformly.

\subsection{Feature Extraction and Fusion} \label{sec:fusion}
Feature extraction is performed on a per-set and per-scene basis, using a
convolutional encoder and hash-grid encodings respectively. The convolutional
encoder is a ResNet-34 \cite{He2016DeepRecognition} truncated after three
stages. The three feature maps are resized to a common resolution and stacked, so every pixel position carries both fine and coarse context. The
stacked map of projection $p$ is its feature map $F_p$, read at
$\pi_p(\mathbf{x})$ by bilinear interpolation \cite{Yu2020PixelNeRF:Images},
which gives one image feature vector
$\mathbf{c}_p = F_p(\pi_p(\mathbf{x}))$ per source projection. Every scene in
the set has its own hash-grid encoding, which stores trainable feature
vectors in tables at several grid resolutions \cite{Muller2022InstantEncoding}.
The active scene's encoding $H_s$ is queried at the point's 3D coordinates by
interpolating the features of the surrounding grid vertices at each
resolution, and the resolutions are concatenated into the vector
$\mathbf{h} = H_s(\mathbf{x})$.

The convolutional encoder is trained by every scene in the set, and
conditioning the model on its image features allows it to learn geometric
priors common to the whole set \cite{Yu2020PixelNeRF:Images}. The hash-grid
encoding serves one scene alone, so $\mathbf{h}$ holds that scene's fine
spatial detail in trainable features rather than network weights, buying
capacity and resolution at the cost of per-scene memory
\cite{Muller2022InstantEncoding}. These two descriptions must be fused before
prediction. The fusion module $G$ concatenates them and applies one learned
linear map,
\begin{equation}
    \mathbf{z}_p = G\left[\, \mathbf{c}_p \, ; \, \mathbf{h} \,\right],
\label{equ:fusion concat}
\end{equation}
producing a fused vector of fixed width $128$ for each source projection.

\subsection{Attenuation Prediction and Aggregation} \label{sec:aggregation}
The prediction head is a fully fused MLP of four layers with $128$ neurons per layer \cite{Muller2021Real-timeTracing}. 
It runs once per source projection, so each point receives $P$ candidate attenuation values $\beta_p$. 
Rendering needs a single value, and the candidates are not equally trustworthy: a point can land outside a projection or near its edge, where the interpolated features are unreliable. 
The candidates are therefore combined by a weighted average in which the model learns how much each projection should count. 
A small linear layer reads $\mathbf{z}_p$ together with two geometric quantities, a flag marking whether the point lands inside projection $p$ and the distance of the landing position from the projection's centre. 
A softmax across the $P$ resulting scores gives the weights, and the point's attenuation is the weighted sum,
\begin{equation}
    \beta = \sum_{p=1}^{P} w_p \, \beta_p , \qquad \sum_{p=1}^{P} w_p = 1 .
\label{equ:view aggregation}
\end{equation}

Multi-view consistency is enforced by construction. 
Every projection is rendered from the same aggregated field, and each update scores all of a scene's target projections at once, so no angle can be fitted at the expense of another.

%% file: contents/Experiments.tex
\section{Results} \label{sec:experiments}

\subsection{Datasets}
Results were obtained on four datasets: two established and two developed
for this comparison. \emph{Ellipsoids} is the synthetic parallel-beam
dataset created by the ONIX authors \cite{Zhang2023ONIX:Views}. It contains
scenes of randomly placed ellipsoids recorded at ten fixed angles over
$0$--$180$\textdegree. \emph{Lung CT} is drawn from LIDC-IDRI
\cite{Armato2011TheScans}, with cone-beam projections generated from the
Computed Tomography (CT) volumes using TIGRE \cite{Biguri2016TIGRE}.

The \emph{Shells} and \emph{Voids} datasets are introduced by this work.
Shells is a synthetic parallel-beam dataset whose scenes enclose a lobed
interior of voids and membranes with far finer structure than Ellipsoids.
Voids is a synthetic cone-beam dataset of volumes containing many small
voids.

Lung CT, Shells and Voids each have a corresponding generator program that can produce
projections at any arbitrary angle from $0$--$180$\textdegree. Ellipsoids is limited
to its ten fixed projections evenly spaced over $0$--$180$\textdegree. Full
specifications for how the datasets were generated are given in the
supplementary material.

\subsection{Implementation and Baseline Methods}
OX-NeRF is programmed in Python \cite{VanRossum2009Python3Manual} and uses
the PyTorch library \cite{Paszke2019PyTorch:Library}. The convolutional
encoder used is the ImageNet pretrained ResNet-34 from torchvision
\cite{He2016DeepRecognition} while the hash encoding and the fully fused
head are from \texttt{tiny-cuda-nn} \cite{Muller2022InstantEncoding,
Muller2021Real-timeTracing}.

\input{tables/hyperparams}

OX-NeRF is compared against four baseline methods that represent the state
of the art for X-ray 3D reconstruction. Each is self-supervised and learns
from projections alone with no volumetric labels. ONIX
\cite{Zhang2023ONIX:Views} is the closest relative of OX-NeRF. It has a
pixel-aligned neural field conditioned on source projections and trained
across scenes. SAX-NeRF \cite{Cai2024Structure-AwareReconstruction} fits one
scene at a time with a line-segment transformer and structure-aware
sampling. R\textsuperscript{2}-Gaussian \cite{Zha2024R2Gaussian} represents
the volume as an explicit set of Gaussians under a rectified radiative model
that keeps the reconstruction tomographically consistent. CombiNeRF
\cite{Bonotto2024CombiNeRF} is a single-scene NeRF that combines several
regularisers for few-shot view synthesis. Two of the four methods required
modification before being compared. CombiNeRF was built for RGB scenes, so
its renderer needed to be replaced by the X-ray transmission model of
Eq.~\ref{equ:path integration}. ONIX, meanwhile, needed to be extended to
support datasets with a cone-beam geometry.

Every method was trained and evaluated on one NVIDIA RTX Pro 6000 GPU.
OX-NeRF used the hyperparameters in Table~\ref{tab:hyperparams} with any
values not listed there following the defaults of
\cite{Muller2022InstantEncoding}. The other methods were trained using their
published default settings. ONIX and OX-NeRF used 50 scenes per dataset for training unless otherwise stated. All other methods were
trained on individual scenes. ONIX and OX-NeRF both condition on source
projections, and in every setting each was given the same four, evenly
spaced across the angular range. Their target projections were all of the
projections available in that setting.

\begin{figure}[t]
    \centering
    \includegraphics[width=\columnwidth]{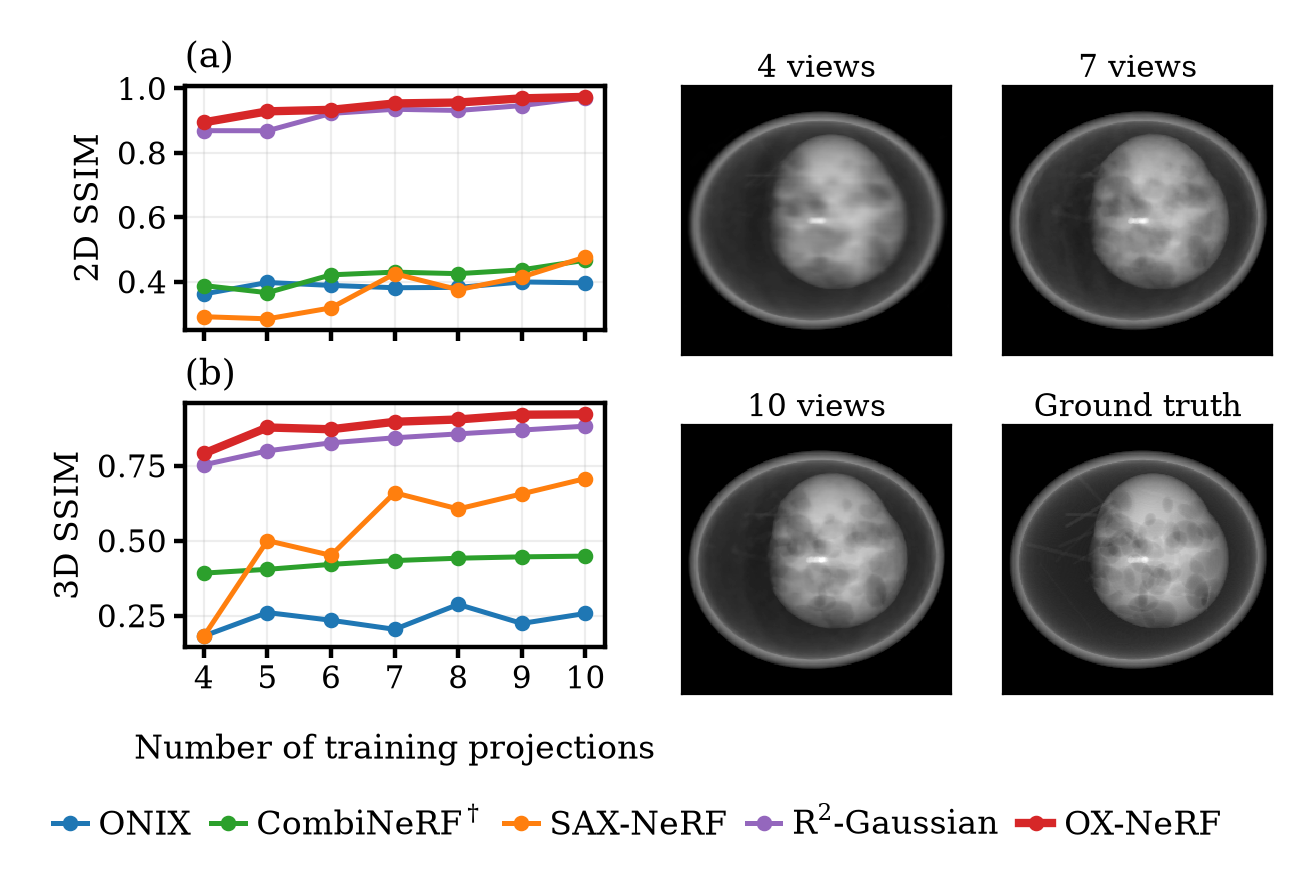}
    \caption{\textbf{Effect of the number of training projections} on the
    Shells dataset. \textbf{(a)} held-out 2D SSIM and \textbf{(b)} 3D SSIM
    as the budget increases from four to ten projections. \textbf{(c)}
    OX-NeRF's prediction of one held-out projection at three budgets on Shells dataset with
    the measurement for reference.}
    \label{fig:views}
\end{figure}

\begin{table*}[t]
    \centering
\caption{\textbf{Novel view synthesis and 3D reconstruction.} NVS reports
    2D SSIM on held-out projections; 3D reports PSNR and SSIM of the
    reconstructed volume against the reference. Best in
    \textcolor{red}{\textbf{red bold}}, second best \underline{underlined}.
    2D PSNR for every cell is in the supplementary material.}

    \setlength{\tabcolsep}{2pt}
    \scriptsize
    \input{tables/main_results}
    \label{tab:main}
\end{table*}

\subsection{Evaluation Setup}
OX-NeRF and the baseline methods were evaluated on two tasks: novel view
synthesis and 3D reconstruction. Novel view synthesis (NVS) asks the model
to render a projection at a held-out angle and scores it against the
measured projection at that angle using 2D SSIM
\cite{Wang2004ImageSimilarity}. 3D reconstruction asks the model for the
full attenuation volume on a voxel grid and scores it against the reference
volume using 3D PSNR and 3D SSIM as defined by R\textsuperscript{2}-Gaussian
\cite{Zha2024R2Gaussian}. The same metric definitions are applied to every
method rather than each method's own evaluation script. Because each method
measures length in its own unit, every reconstructed volume is the true
volume up to one positive constant, which a single least-squares gain
removes before scoring. Full definitions of metrics and methodology are
provided in the supplementary material.

Table~\ref{tab:main} reports both tasks at three view budgets per dataset.
Ellipsoids is tested up to nine views because at least one of its ten fixed
angles must be held out. On the other three datasets the 2D SSIM is averaged
over five held-out angles.

\begin{figure*}[t]
    \centering
    \includegraphics[width=0.78\textwidth]{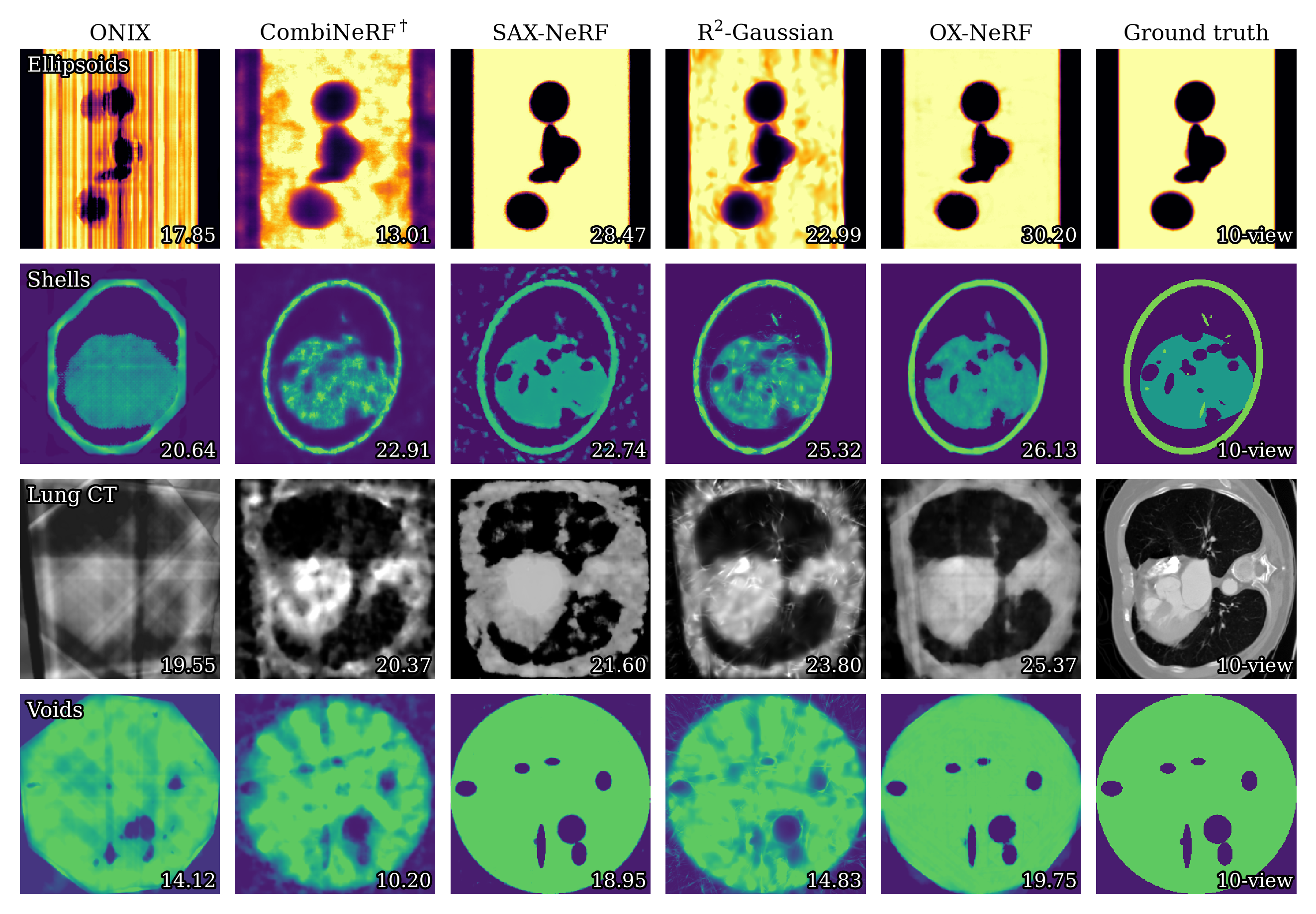}
    \caption{\textbf{Qualitative 3D reconstruction} at ten training views. Each panel is
    rendered from the reconstructed volume under an identical display window taken from the
    ground truth, so panels are directly comparable; the inset number is 3D PSNR in dB.}
    \label{fig:ct_visual}
\end{figure*}

\subsection{Novel View Synthesis and 3D Reconstruction}
OX-NeRF renders the most faithful held-out projections in eleven of the
twelve settings described in Table~\ref{tab:main}. The separation is
clearest in the settings with fewer views. Figure~\ref{fig:views} traces
the evolution from four to ten views. As the number of views increases the
reconstruction quality of every method improves, as expected, and the
advantage of OX-NeRF in NVS becomes less pronounced. By ten views the
advantage is marginal and nearly every method renders excellent
projections. A rendered projection can, however, agree closely with its
measurement without the volume behind it being correct.


On the reconstructed volumes the advantage of OX-NeRF is unambiguous. It
achieves the highest 3D SSIM in every setting and the highest 3D PSNR in
all but one. The gap between the two tasks is easiest to see at ten
views, where the 2D SSIM scores of most methods were high and close
together. Figure~\ref{fig:ct_visual} visualises a slice through the
reconstructed volume of each method at that budget. The baseline methods
capture the coarse structure but have significant artefacts. These include
streaks aligned with the projection directions, a signature of too few
angles, and a diffuse haze that fills regions that should be empty. The
OX-NeRF reconstructions demonstrate finer detail and have significantly
fewer artefacts.

\input{tables/ablation_row}

\subsection{Ablation} \label{sec:ablation}
Three design choices define OX-NeRF: the per-scene hash grid, the
convolutional encoder and residual-guided ray selection. Panel (a) of
Table~\ref{tab:ablation} demonstrates the effect of these choices using the
Ellipsoids dataset at six training views. The hash grid is compared against
the fixed sinusoidal positional encoding of NeRF
\cite{Mildenhall2021NeRF:Synthesis, Tancik2020FourierDomains}, which maps
each coordinate through sines and cosines of increasing frequency and is
the standard point encoder that the hash grid was introduced to supersede. The convolutional encoder is compared against its absence, which reduces
the model to a per-scene neural field with no information from the other
scenes in the set. Residual-guided selection is compared against
gradient-guided selection \cite{SobelFeldman1968Gradient,
Gao2010AnDetection}, which scores each pixel once by its Sobel image
gradient and draws rays in proportion. The gradient approach is a natural
alternative as both selection methods try to concentrate rays where
information is expected to be heuristically.

The convolutional encoder improves on either point encoder alone, and
paired with the hash grid gives the best configuration by a wide margin.
The hash grid outperforms the sinusoidal encoding in every configuration
but one: with gradient-guided rays and no encoder it gives the weakest 3D
SSIM in the panel, since without a prior or error feedback its additional
capacity has nothing to constrain it. Residual-guided selection improves
3D SSIM in every configuration, and its effect is largest on that same
unconstrained hash grid. The gradient rule scores the measurements rather
than the model, so it keeps sending rays to edges that are already
reconstructed, whereas the residual rule follows the error as it moves.
Taken together, the cross-scene encoder and the per-scene hash grid account
for most of the improvement and are strongest in combination, with residual
sampling providing a further boost.

Panels (b) and (c) of Table~\ref{tab:ablation} vary the number of scenes in
the set and the hash table size to test whether the shared prior benefits
from more scenes and the per-scene representation from more capacity.
Reconstruction quality rises sharply as scenes are added up to 50 and more
gradually beyond, and it rises with table size at every step up to
$\log_2 T = 14$ before levelling off. The trade-off is GPU memory. Every
scene keeps its own hash table resident throughout training. Since each
increment of $\log_2 T$ doubles the size of every table the memory
consumed by the hash grids scales with the product of the two. Notably, the
defaults of 50 scenes and $\log_2 T = 12$ are not the strongest settings in
the ablation. It is instead the largest that this memory trade-off allows
on every dataset with Lung CT being the limiting factor.
\input{tables/cb1d_high_views}

\subsection{Increasing Views}
OX-NeRF is designed for the ultra-sparse domain. It can be applied to more
than ten views but as the number of views increases both theoretical
and technical issues arise. The convolutional encoder accounts for a significant portion of the
compute budget supplying a prior between angles. The value of that
prior, however, falls as the gaps between angles close. Lung CT, the hardest of the
four datasets, makes a useful test case. Table~\ref{tab:cb1d_high_views}
extends the comparison against R\textsuperscript{2}-Gaussian, the strongest
baseline, to 15, 20 and 25 views. OX-NeRF still leads at 15. At 20 the two
methods are on par, and by 25 R\textsuperscript{2}-Gaussian has clearly
overtaken. The crossover matches the two designs:
R\textsuperscript{2}-Gaussian spends its compute on rendering more rays and
as views are added that becomes a better strategy. The OX-NeRF results
could likely be improved by increasing the number of rays per iteration, the
number of scenes in the set, the hash table size or the number of source
projections.
Each of these costs memory, however, and with the chosen experimental settings
OX-NeRF already peaks at over 90\,GB of VRAM.

%% file: tables/hyperparams.tex
\begin{table}[t]
    \centering
    \caption{\textbf{Hyperparameters.} One recipe is used for all datasets.
    $P$ is the number of source projections; $\tau$ and $B$ are the
    residual-map refresh period and scene count of
    Section~\ref{sec:sampling}; $T$ is the hash table size.}
    \label{tab:hyperparams}
    \footnotesize
    \setlength{\tabcolsep}{4pt}
    \begin{tabular}{@{}lp{0.66\columnwidth}@{}}
        \toprule
        Component & Setting \\
        \midrule
        Hash grid & Per-scene table; 16 levels; 8 features per level; base resolution 16, growth factor 1.5; $\log_2 T = 12$ \\
        Image encoder & ResNet-34, first three stages; ImageNet initialisation; $P = 4$ source projections \\
        Rays & 2048 rays per iteration; 320 points sampled per ray \\
        Residual sampling & $\alpha = 1$; $\varepsilon = 0.25$; refresh every $\tau = 5$ epochs over $B = 8$ scenes \\
        Optimisation & 20\,000 iterations; Adam, lr $5\times10^{-4}$ ($2.5\times10^{-4}$ for fusion); momentum 0.9, 0.99 \cite{Muller2022InstantEncoding} \\
        \bottomrule
    \end{tabular}
\end{table}

%% file: tables/main_results.tex

\begin{tabular}{lcccccccccccccccc}
\toprule
\multicolumn{2}{c}{} & \multicolumn{3}{c}{ONIX~\cite{Zhang2023ONIX:Views}} & \multicolumn{3}{c}{CombiNeRF~\cite{Bonotto2024CombiNeRF}} & \multicolumn{3}{c}{SAX-NeRF~\cite{Cai2024Structure-AwareReconstruction}} & \multicolumn{3}{c}{R$^2$-Gaussian~\cite{Zha2024R2Gaussian}} & \multicolumn{3}{c}{\textbf{OX-NeRF}} \\
\cmidrule(lr){3-5}\cmidrule(lr){6-8}\cmidrule(lr){9-11}\cmidrule(lr){12-14}\cmidrule(lr){15-17}
\multicolumn{2}{c}{} & NVS & \multicolumn{2}{c}{3D} & NVS & \multicolumn{2}{c}{3D} & NVS & \multicolumn{2}{c}{3D} & NVS & \multicolumn{2}{c}{3D} & NVS & \multicolumn{2}{c}{3D} \\
\cmidrule(lr){4-5}\cmidrule(lr){7-8}\cmidrule(lr){10-11}\cmidrule(lr){13-14}\cmidrule(lr){16-17}
Dataset & Views & SSIM & PSNR & SSIM & SSIM & PSNR & SSIM & SSIM & PSNR & SSIM & SSIM & PSNR & SSIM & SSIM & PSNR & SSIM \\
\midrule
\multirow{3}{*}{Ellipsoids} & 4 & 0.7466 & \underline{17.33} & 0.3331 & 0.6494 & 12.14 & 0.2434 & 0.6406 & 12.36 & \underline{0.7037} & \underline{0.8069} & 16.40 & 0.6725 & \textcolor{red}{\textbf{0.8893}} & \textcolor{red}{\textbf{19.28}} & \textcolor{red}{\textbf{0.8406}} \\
 & 7 & 0.7843 & 20.45 & 0.3161 & 0.6247 & 12.38 & 0.2508 & \underline{0.9394} & \underline{27.31} & \underline{0.9415} & 0.8364 & 19.16 & 0.7279 & \textcolor{red}{\textbf{0.9915}} & \textcolor{red}{\textbf{27.89}} & \textcolor{red}{\textbf{0.9566}} \\
 & 9 & 0.8098 & 20.21 & 0.2518 & 0.7360 & 13.01 & 0.2555 & 0.9620 & \textcolor{red}{\textbf{28.84}} & \underline{0.9464} & \textcolor{red}{\textbf{0.9996}} & 22.67 & 0.7719 & \underline{0.9966} & \underline{28.55} & \textcolor{red}{\textbf{0.9613}} \\
\midrule
\multirow{3}{*}{Shells} & 4 & 0.3621 & 17.12 & 0.1822 & 0.3883 & 18.31 & 0.3921 & 0.2916 & 16.98 & 0.1816 & \underline{0.8689} & \underline{18.87} & \underline{0.7533} & \textcolor{red}{\textbf{0.8947}} & \textcolor{red}{\textbf{20.77}} & \textcolor{red}{\textbf{0.7912}} \\
 & 7 & 0.3815 & 20.30 & 0.2046 & 0.4299 & 21.59 & 0.4339 & 0.4249 & 20.06 & 0.6601 & \underline{0.9353} & \underline{23.21} & \underline{0.8435} & \textcolor{red}{\textbf{0.9534}} & \textcolor{red}{\textbf{24.45}} & \textcolor{red}{\textbf{0.8969}} \\
 & 10 & 0.3966 & 20.64 & 0.2576 & 0.4669 & 22.91 & 0.4488 & 0.4767 & 22.74 & 0.7076 & \underline{0.9718} & \underline{25.32} & \underline{0.8824} & \textcolor{red}{\textbf{0.9738}} & \textcolor{red}{\textbf{26.13}} & \textcolor{red}{\textbf{0.9220}} \\
\midrule
\multirow{3}{*}{Lung CT} & 4 & 0.8794 & 19.18 & 0.5416 & 0.8995 & 18.45 & 0.5417 & 0.9044 & 18.82 & \underline{0.5657} & \underline{0.9205} & \underline{20.18} & 0.5507 & \textcolor{red}{\textbf{0.9399}} & \textcolor{red}{\textbf{21.58}} & \textcolor{red}{\textbf{0.6235}} \\
 & 7 & 0.9192 & 19.71 & 0.5583 & 0.9004 & 19.32 & 0.5404 & 0.9272 & 20.75 & \underline{0.6050} & \underline{0.9346} & \underline{22.36} & 0.6024 & \textcolor{red}{\textbf{0.9525}} & \textcolor{red}{\textbf{23.77}} & \textcolor{red}{\textbf{0.6769}} \\
 & 10 & 0.9153 & 19.55 & 0.5478 & 0.9150 & 20.37 & 0.5510 & 0.9451 & 21.60 & 0.5694 & \underline{0.9617} & \underline{23.80} & \underline{0.6423} & \textcolor{red}{\textbf{0.9667}} & \textcolor{red}{\textbf{25.37}} & \textcolor{red}{\textbf{0.7102}} \\
\midrule
\multirow{3}{*}{Voids} & 4 & \underline{0.9537} & \underline{13.00} & 0.5004 & 0.9271 & 9.22 & 0.3341 & 0.9535 & 12.54 & \underline{0.5583} & 0.9465 & 11.83 & 0.3471 & \textcolor{red}{\textbf{0.9579}} & \textcolor{red}{\textbf{13.71}} & \textcolor{red}{\textbf{0.5938}} \\
 & 7 & 0.9438 & 13.73 & 0.6181 & 0.9338 & 9.52 & 0.3483 & \underline{0.9789} & \underline{18.26} & \underline{0.8132} & 0.9606 & 13.82 & 0.3751 & \textcolor{red}{\textbf{0.9853}} & \textcolor{red}{\textbf{19.61}} & \textcolor{red}{\textbf{0.8434}} \\
 & 10 & 0.9545 & 14.12 & 0.5438 & 0.9451 & 10.20 & 0.3621 & \underline{0.9814} & \underline{18.95} & \underline{0.8552} & 0.9721 & 14.83 & 0.4025 & \textcolor{red}{\textbf{0.9863}} & \textcolor{red}{\textbf{19.75}} & \textcolor{red}{\textbf{0.8677}} \\
\bottomrule
\end{tabular}

%% file: tables/ablation_row.tex
\begin{table}[t]
    \centering
    \caption{\textbf{Ablations on Ellipsoids at six target projections.}
        Panel~(a) is a full factorial over the point encoder (sinusoidal or
        hash grid), the convolutional image encoder (present or absent) and
        the ray-selection rule ($\nabla$ gradient-guided, $R$ residual-guided).
        Panels~(b) and~(c) use the full model and vary the number of scenes in
        the set and the hash table size $T$. All runs share one recipe and one
        held-out view. Metrics are held-out 2D SSIM and 3D SSIM. Best in
        \textbf{bold}, second best \underline{underlined}.}
    \label{tab:ablation}
    \setlength{\tabcolsep}{3pt}
    \renewcommand{\arraystretch}{0.98}
    \scriptsize
    \textbf{(a) Model components}\\[2pt]
    \input{tables/ablation_components}\\[8pt]
    \adjustbox{valign=t}{\begin{tabular}[t]{@{}c@{}}
        \textbf{(b) Training scenes}\\[2pt]
        \input{tables/ablation_scenes_residual}
    \end{tabular}}%
    \hspace{1.5em}%
    \adjustbox{valign=t}{\begin{tabular}[t]{@{}c@{}}
        \textbf{(c) Hash capacity}\\[2pt]
        \input{tables/ablation_capacity_residual}
    \end{tabular}}%
\end{table}

%% file: tables/ablation_components.tex
\begin{tabular}{@{}clccc@{}}
\toprule
Rays & Point encoder & Image encoder & 2D SSIM & 3D SSIM \\
\midrule
$\nabla$ & Sinusoidal &  & 0.6481 & 0.3576 \\
$R$ & Sinusoidal &  & 0.8011 & 0.4431 \\
$\nabla$ & Sinusoidal & \checkmark & 0.8745 & 0.7913 \\
$R$ & Sinusoidal & \checkmark & 0.9141 & 0.8135 \\
$\nabla$ & Hash grid &  & 0.6521 & 0.2797 \\
$R$ & Hash grid &  & \underline{0.9553} & 0.6414 \\
$\nabla$ & Hash grid & \checkmark & 0.9450 & \underline{0.8906} \\
$R$ & Hash grid & \checkmark & \textbf{0.9858} & \textbf{0.9542} \\
\bottomrule
\end{tabular}

%% file: tables/ablation_scenes_residual.tex
\begin{tabular}{@{}ccc@{}}
\toprule
Scenes & 2D SSIM & 3D SSIM \\
\midrule
25 & 0.9569 & 0.9063 \\
50 & 0.9794 & \underline{0.9450} \\
75 & \underline{0.9828} & 0.9446 \\
100 & \textbf{0.9859} & \textbf{0.9455} \\
\bottomrule
\end{tabular}

%% file: tables/ablation_capacity_residual.tex
\begin{tabular}{@{}ccc@{}}
\toprule
$\log_2 T$ & 2D SSIM & 3D SSIM \\
\midrule
10 & 0.9539 & 0.8784 \\
12 & 0.9678 & 0.9062 \\
14 & \textbf{0.9823} & \underline{0.9418} \\
16 & \underline{0.9781} & \textbf{0.9420} \\
\bottomrule
\end{tabular}

%% file: tables/cb1d_high_views.tex
\begin{table}[t]
    \centering
    \caption{\textbf{High-view Lung CT stress test.} The dense Lung CT scene uses
    the same cone geometry as Table~\ref{tab:main} but extends the training budget to
    15, 20 and 25 projections. Columns report 3D reconstruction scores as in
    Table~\ref{tab:main}. Best in \textcolor{red}{\textbf{red bold}}.}
    \label{tab:cb1d_high_views}
    \scriptsize
    \setlength{\tabcolsep}{4pt}
    \begin{tabular}{@{}ccccc@{}}
        \toprule
        \multicolumn{1}{c}{} & \multicolumn{2}{c}{\textbf{OX-NeRF}} & \multicolumn{2}{c}{R$^2$-Gaussian} \\
        \cmidrule(lr){2-3}\cmidrule(lr){4-5}
        \textbf{Views} & 3D PSNR & 3D SSIM & 3D PSNR & 3D SSIM \\
        \midrule
        15 & \textcolor{red}{\textbf{26.47}} & \textcolor{red}{\textbf{0.7442}} & {25.65} & {0.6984} \\
        20 & \textcolor{red}{\textbf{27.04}} & {0.7493} & {26.97} & \textcolor{red}{\textbf{0.7500}} \\
        25 & {27.14} & {0.7537} & \textcolor{red}{\textbf{27.91}} & \textcolor{red}{\textbf{0.7871}} \\
        \bottomrule
    \end{tabular}
\end{table}

%% file: contents/Conclusions.tex
\section{Conclusions} \label{sec:conclusions}
OX-NeRF combines a prior learned across a set of scenes with a spatial
representation owned by each scene. In the ultra-sparse domain of ten
projections or fewer, it demonstrates a significant improvement in 3D
reconstruction accuracy over existing self-supervised methods. The result
holds on parallel-beam and cone-beam data alike.

There are several potential changes to be investigated that could improve
reconstruction accuracy. The most immediate follows from the ablation: the
number of scenes and the hash table size both raise reconstruction quality
and both were capped by memory, so whether the gains continue on more
capable hardware is an open question.
Domain-specific priors are another possibility, since building known
anatomical structure into a reconstruction has proven effective
\cite{Yu2025X2GaussianReconstruction} and the shared components of OX-NeRF
could in principle host priors of that kind. Cross-scene Gaussian splatting
for X-ray is another route to explore. Cross-scene priors for splatting exist in
visible-light imaging \cite{Charatan2024PixelSplat:Reconstruction} and
splatting is already established for tomographic reconstruction
\cite{Zha2024R2Gaussian}, but the combination has not yet been attempted.

The most exciting future direction is extending OX-NeRF into the time
domain so it can be applied to high-rate acquisition. The ultra-sparse
regime arises in this setting by necessity as a source that captures a
transient event in a single shot can record only as many projections as it
has simultaneous views. One example is X-ray multi-projection imaging at
synchrotrons and free-electron lasers, which records a handful of
projections per frame at kilohertz to megahertz frame rates
\cite{Vagovic2018HardApproaches, Villanueva-Perez2023MegahertzImaging,
Asimakopoulou2024DevelopmentImaging, Rosen2024SynchrotronFlows,
Witte20264DImaging}. Every frame shares one acquisition and one geometry.
The problem reframes naturally: the input changes from a set to a temporally ordered sequence, and nothing else in the formulation has to
change. Further, a sequence could provide useful temporal regularisation given that geometry which is ambiguous in one frame of the sequence could be resolved in another as the objects within it move. Time-resolved reconstruction of simple scenes from sparse projections has already been shown to be feasible \cite{Zhang20254D-ONIXLearning, Yue2025Physics-informedData}. The aim is to build on this and use OX-NeRF to reconstruct more complex scenes that demonstrate high-rate phenomena.

%% file: back/Acknowledgement.tex
\section*{Acknowledgements}
We thank Daniel Moško for discussions on the ONIX method, Yi-chen Ju for ideas on the fusion scheme, and Yuhe Zhang for sharing the ellipsoids dataset. This work was supported by the United States Air Force Office of Scientific Research under award number FA8655-24-1-7353, and by the EPSRC and First Light Fusion under the AMPLIFI Prosperity Partnership (EP/X025373/1), funding the NVIDIA Blackwell GPU used in this research. Any opinions, findings, and conclusions or recommendations expressed in this material are those of the author(s) and do not necessarily reflect the views of the United States Air Force.